%% file: colm2025_conference.tex
\documentclass{article} 
\usepackage[preprint]{colm2025_conference}

\usepackage{microtype}
\usepackage{hyperref}
\usepackage{url}
\usepackage{booktabs}
\usepackage{amsmath}
\usepackage{yfonts}
\usepackage{graphicx}
\usepackage{wrapfig}

\usepackage{yfonts}     
\usepackage{microtype}  
\usepackage{pifont}     
\usepackage{lineno}
\usepackage[frozencache,cachedir=minted-cache]{minted}
\usepackage{amsmath,amssymb}
\usepackage{csquotes}

\definecolor{darkblue}{rgb}{0, 0, 0.5}
\hypersetup{colorlinks=true, citecolor=darkblue, linkcolor=darkblue, urlcolor=darkblue}

\def\Snospace~{\S{}}

\definecolor{coco1}{HTML}{D9E4EC}
\definecolor{coco2}{HTML}{B7CFDC}
\definecolor{coco3}{HTML}{6AABD2}
\definecolor{coco4}{HTML}{385E72}
\hypersetup{
    colorlinks=true,
    linkcolor=coco3,
    filecolor=coco4,      
    urlcolor=coco3,
    citecolor=coco3,
}

\title{Learning to Simulate Human Dialogue}

\author{
Kanishk Gandhi\footnotemark[1]\\
Stanford University
\And
Agam Bhatia\thanks{Equal Contribution}\\
Stanford University
\And
Noah D. Goodman
\\
Stanford University
}

\begin{document}

\ifcolmsubmission
\linenumbers
\fi

\maketitle

\begin{abstract}
\vspace{-2mm}
\input{sections/00_abstract}
\end{abstract}
\vspace{-10mm}
\section{Introduction}

\input{sections/01_introduction}

\section{Related Work}
\input{sections/02_related}

\begin{figure}
    \centering
    \includegraphics[width=0.95\linewidth]{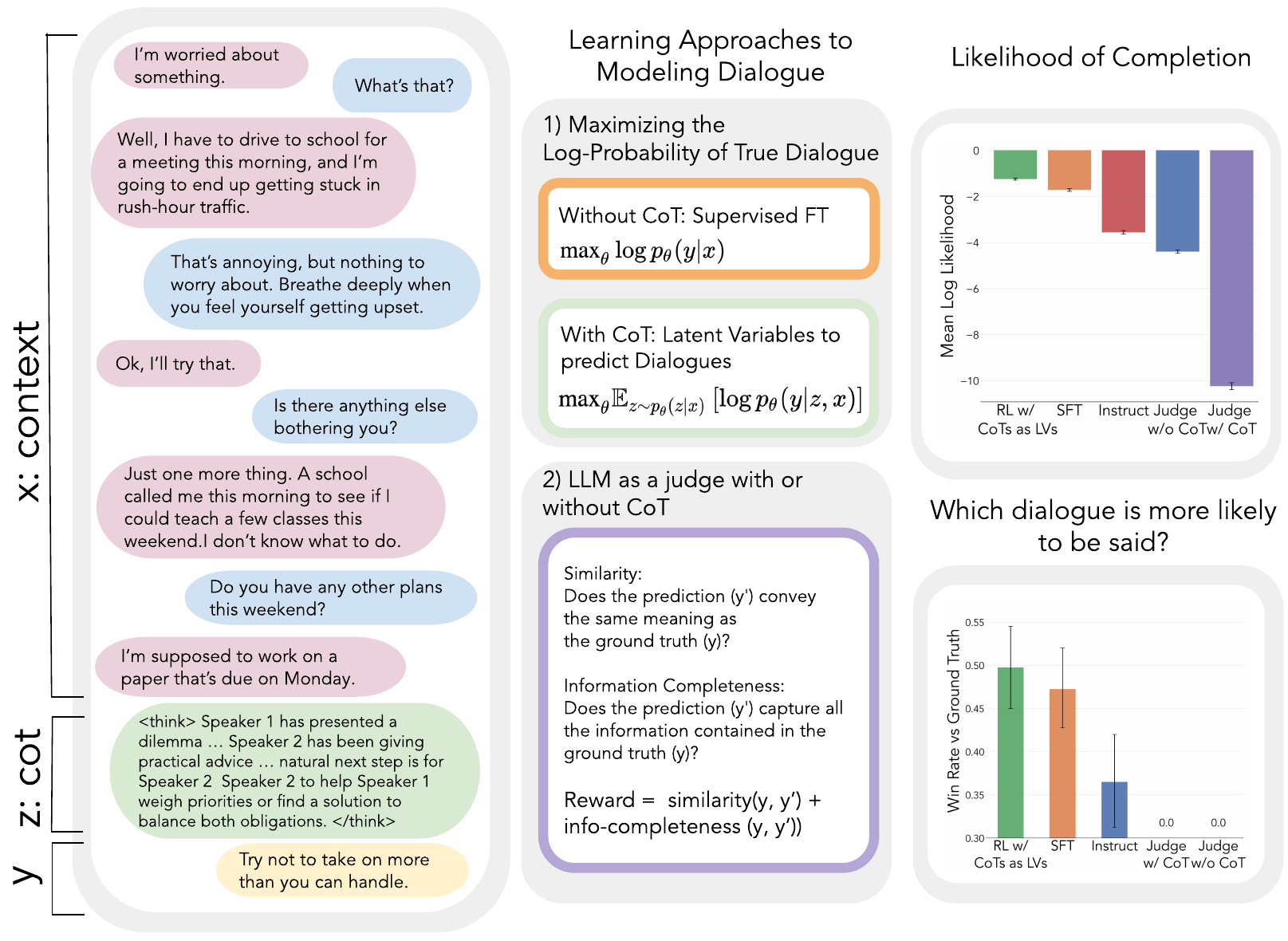}
    \vspace{-3mm}
    \caption{
    \textit{(left)} \textbf{Learning to simulate human dialogue.} Predicting the next dialogue
$y$ given context $x$, optionally reasoning via chain-of-thought $z$. \textit{(center)} We vary the reward signal (LLM-as-a-judge vs. log-probability) and reasoning mode (thinking vs. no thinking). With thinking, chain-of-thought serves as a latent variable optimized to increase likelihood of true human responses. \textit{(right)} Log-probability training yields higher ground-truth likelihood on the test set and human win rates; judge-based training leads to reward hacking. Thinking amplifies both effects.}
    \label{fig:front}
\end{figure}
\vspace{-2mm}
\section{Learning Approaches to Modeling Dialogue}
\vspace{-2mm}
\input{sections/03_method}

\section{Experiments}
\input{sections/04_results}
\section{Discussion}

\input{sections/05_discussion}

\section*{Acknowledgments}
We would like to thank Violet Xiang, Omar Shaikh, Eric Zelikman, Jan Philipp-Fr\"anken and Rose Wang for discussions.
KG was supported by an HAI-SAP Grant and
an NSF Expeditions grant.

\bibliography{colm2025_conference}
\bibliographystyle{colm2025_conference}
\clearpage

\appendix
\input{sections/10_appendix}

\end{document}

%% file: sections/00_abstract.tex
To predict what someone will say is to model how they think. We study this through next-turn dialogue prediction: given a conversation, predict the next utterance produced by a person. We compare learning approaches along two dimensions: (1) whether the model is allowed to think before responding, and (2) how learning is rewarded—either through an LLM-as-a-judge that scores semantic similarity and information completeness relative to the ground-truth response, or by directly maximizing the log-probability of the true human dialogue. 
We find that optimizing for judge-based rewards indeed increases judge scores throughout training, however it decreases the likelihood assigned to ground-truth human responses and decreases the win rate when human judges choose the most human-like response among a real and synthetic option.
This failure is amplified when the model is allowed to think before answering. 
In contrast, by directly maximizing the log-probability of observed human responses, the model learns to better predict what people actually say, improving on both log-probability and win rate evaluations.
Treating chain-of-thought as a latent variable, we derive a lower bound on the log-probability.
Optimizing this objective yields the best results on all our evaluations. 
These results suggest that thinking helps primarily when trained with a distribution-matching objective grounded in real human dialogue, and that scaling this approach to broader conversational data may produce models with a more nuanced understanding of human behavior.\footnote{Code available at: \href{htts://github.com/kanishkg/dialogue-sim}{https://github.com/kanishkg/dialogue-sim}}

\begin{flushright}
``Man is a mystery. It needs to be unravelled, and if you spend your whole life unravelling it, don't say you've wasted time. I am studying that mystery because I want to be a human being.'' \\---Dostoevsky
\end{flushright}

%% file: sections/01_introduction.tex
How can we build language models that understand human behavior? We propose a simple method: train them to predict what people will say. A model that can accurately anticipate the next turn in a human conversation must, implicitly, reason about the speaker's intent, beliefs, and social context. This capacity for human simulation becomes increasingly important as LLMs are deployed in settings requiring sustained interaction, from long horizon dialogue agents to systems that learn directly from real user conversations \citep{weston2025ai}.

Pretraining may already imbue language models with some capacity for simulating humans --- after all, the text in the pretraining data is written by humans with an intent to communicate, inherent with some belief about the world \citep{andreas2022language} and learning to predict the next token written by a person implicitly optimizes an LLM to understand and generate human-like utterances. Yet, this capacity remains inconsistent: while language models show human-like patterns on some social reasoning tasks, they fail on trivial alterations and struggle to generalize across contexts \citep{gandhi2023understanding,gandhi2024human,shapira2024clever,ullman2023large}. One response has been to build elaborate inference-time scaffolds that layer memory, persona descriptions, and multi-step reasoning on top of base models \citep{park2023generative,hewitt2024predicting}. These agentic systems can produce compelling human-like behavior, but their capabilities come from human engineering with limited ability to learn from behavioral data. We ask what the right training methods might be to make models better at predicting human behavior.

We study this question through the task of next turn dialogue prediction: given a conversation, predict what a person will say next (\autoref{fig:front}). We explore different learning approaches, with two reward types: first, an LLM-as-a-judge approach that compares a generated response to the ground-truth human utterance in terms of similarity and information completeness; second, the log-probability assigned by the model to the true human response.
In both cases, we also vary whether the model is allowed to think before responding. The log-probability cannot be directly optimized when thinking is allowed, since tokens sampled by the model are discrete and cannot be backpropagated through. We can, however, treat the chain-of-thought as a discrete latent variable \citep{hoffman2023training} and derive a lower bound (the evidence lower bound, ELBO, of variational inference methods).
This rewards thoughts for increasing the likelihood of the true human response and increases the likelihood of that response given the thoughts.

Our experiments reveal that training with LLM-as-a-judge rewards does not improve human dialogue prediction. Models trained with an LLM judge achieve increasing reward scores throughout training, yet assign lower likelihood to actual human responses. Despite extensive efforts to use different rubrics to train, models usually end up ``reward hacking'', failing to capture the true distribution. This reward hacking is amplified when models are allowed to think before responding. 

In contrast, directly optimizing for the log-probability of true human responses is more effective. Supervised finetuning on the true dialogues gives more human-like responses. 
Training with chain-of-thought as a latent variable works better: by directly optimizing for the log probability of true human responses, with reasoning rewarded only if it helps predict what humans say, these models exceed supervised finetuning. 
We validate these findings through a blind human preference study on next-turn dialogue prediction. Given a conversation context, annotators chose which completion (model generated or ground-truth human dialogue) was more appropriate. Human evaluators consistently prefer responses from the log-probability trained models, with the latent variable approach achieving the highest win rates. 

Our results suggest that a simple distributional objective can help build better models of human dialogue, especially when combined with chain-of-thought. Scaling this approach to larger and more diverse conversation data may yield models with a more nuanced understanding of human behavior.

%% file: sections/02_related.tex
Our work draws on three complementary lines of research: agentic approaches to modeling human behavior, learning-based methods for human simulation, and RL in non-verifiable domains.

\textbf{Agentic Scaffolds for Human Simulation.} One approach to modeling human behavior is to use carefully designed inference-time scaffolds that incorporate memory, reasoning, and individual characteristics drawn from the observations of a person \citep{park2023generative,shaikh2025creating,sumers2023cognitive,hewitt2024predicting}. These systems operate at inference time, where human-like behavior emerges from prompt engineering and the orchestration of multiple LLM calls rather than from direct learning on behavioral data. Our work takes a complementary approach: we train language models to predict human dialogue directly, encouraging them to internalize social reasoning in their weights. We view agentic scaffolds as complementary to our method and training such systems using similar objectives could further improve human simulation.

\textbf{Learning to Simulate Humans.} Prior work on training dialogue agents to model individual users has largely optimized for proxies of human simulation like persona consistency \citep{zhang2018personalizing,abdulhai2025consistently}, user modeling accuracy \citep{wan2025enhancing}, or preference elicitation \citep{andukuri2024star}. We instead focus on directly predicting actual human behavior. This objective requires implicit modeling of intent, beliefs, and social context, capacities that we argue are essential for effective human simulation.

\textbf{Social Reasoning in Language Models.} Human behavior, including dialogue, can often be explained in terms of hidden mental states: a capacity commonly referred to as theory of mind \citep{premack1978does,lake2017building}. Studies have shown that while LLMs can model beliefs, desires, and affect in human-like ways \citep{kosinski2024evaluating,gandhi2023understanding,gandhi2024human}, this ability remains inconsistent across contexts \citep{shapira2024clever,ullman2023large,kim2023fantom}. While the capacity for social reasoning might emerge through pretraining alone \citep{andreas2022language}, we show how  explicitly training to predict future human behavior might lead better capabilities.

\textbf{Reinforcement Learning Beyond Verifiable Domains.} Using RL in domains without ground truth answers, such as dialogue, creative writing, and open-ended reasoning, requires alternative reward formulations. One approach uses LLM judges to evaluate responses based on specified criteria \citep{zheng2023judging}, enabling scalable approximation of human preferences. A second approach bypasses judges by treating chain-of-thought reasoning as a latent variable \citep{hoffman2023training,zelikman2024quiet,gurung2025learning,hatamizadeh2025rlp}. The intuition behind these approaches is to reward reasoning based on increasing the likelihood of the observed data, with no external verifier or judge. We explore both approaches in the context of human dialogue prediction.

%% file: sections/03_method.tex
\paragraph{Task Setup.} We study the task of next-turn dialogue prediction: given a conversational context with two speakers, $x = (u_1, u_2, \ldots, u_{t-1})$ consisting of alternating speaker turns (see \autoref{fig:front}), the goal is to generate the next human utterance $y=u_t$. This task requires models to capture intent, maintain coherence, and produce responses that are semantically aligned with ground-truth human speech. Unlike open-ended dialogue generation where multiple valid responses exist, we evaluate against specific human-produced continuations. 

To create our training data, we perform data augmentation by splitting each multi-turn dialogue at every intermediate turn. Specifically, for a dialogue with $n$ turns, we generate $n-1$ examples where the $i$-th example uses turns $\{u_1, u_2, \ldots, u_i\}$ as context to predict turn $u_{i+1}$. 

We explore learning approaches along two dimensions: (1) the reward signal—either an LLM-as-a-judge or the log-probability of ground-truth human responses, and (2) whether the model generates chain-of-thought reasoning before responding, yielding 4 conditions. For all experiments we use Qwen-2.5-3B-Instruct as the model being trained.\footnote{We chose an instruct model over base as it followed the formatting instructions more consistently.}

\paragraph{Reinforcement Learning with LLM-as-a-Judge.} We train models using Group Relative Policy Optimization (GRPO; \citet{shao2024deepseekmath,yu2025dapo,liu2025understanding}) with rewards provided by an LLM judge. For each context, we sample $G=16$ candidate responses and compute rewards based on two criteria evaluated by Qwen-2.5-3B-Instruct: (1) \textit{semantic similarity} to the ground-truth response, measuring whether the generated response captures the same meaning and intent, and (2) \textit{information completeness}, measuring whether all relevant information from the ground-truth is preserved. We explored additional reward dimensions including intentionality and stylistic alignment, but found these showed poor agreement with human raters and thus excluded them from the final training objective (see \autoref{sec:judge} for a detailed discussion). The total reward is the sum of semantic similarity and information completeness scores, each normalized to $[0, 1]$. We use a low KL coefficient ($\beta = 0.001$) and mask truncated completions during advantage estimation, as in \citet{yu2025dapo}. We investigate two variants of this approach: \textit{1) With Chain-of-Thought}: The model generates an explicit reasoning trace before producing its final response, enclosed in \texttt{<think>...</think>} tags. This allows the model to deliberate about conversational context and potential responses. \textit{2) Without Chain-of-Thought}: The model directly generates responses without intermediate reasoning steps.

\paragraph{Optimizing for the Log-Probabilities.} Rather than relying on a judge, we directly optimize for the log-probability of the ground-truth human response given the context. This distribution-matching objective provides a principled training signal grounded in actual human behavior, avoiding the potential for reward hacking inherent in proxy rewards. We investigate two variants: \textit{1) Without Chain-of-Thought}: The model is finetuned to directly predict the next human turn given the conversational context using standard causal language modeling loss, equivalent to supervised fine-tuning on (dialogue context, response) pairs. \textit{2) With Chain-of-Thought}: Motivated by approaches that avoid explicit judge models \citep{hoffman2023training,zelikman2024quiet, zhou2025reinforcing,hatamizadeh2025rlp,gurung2025learning}, we treat the chain-of-thought as a latent variable that helps predict human responses. Given conversation context $x$ and ground-truth human response $y$, we introduce thoughts $z$ and model the marginal likelihood:

\begin{equation}
p(y|x) = \int p(y|x,z) \, p(z|x) \, dz
\end{equation}

Directly optimizing the marginal likelihood is unstable; we instead follow the usual derivation of the evidence lower bound on log- marginal probability, using Jensen's inequality:
\begin{equation}
\log p(y|x) = \log \mathbb{E}_{z \sim p_\theta(z|x)}[p_\theta(y|x,z)] \geq \mathbb{E}_{z \sim p_\theta(z|x)}[\log p_\theta(y|x,z)],
\end{equation}
leading us to the objective:
\begin{equation}
\mathcal{L}(\theta) = \mathbb{E}_{z \sim p_\theta(z|x)}[\log p_\theta(y|x,z)].
\end{equation}
To derive the gradient of this objective we expand the expectation, apply the product rule, and use the log-derivative trick ($\nabla_\theta p_\theta(z|x) = p_\theta(z|x) \nabla_\theta \log p_\theta(z|x)$):
\begin{align}
\nabla_\theta \mathcal{L} &= \nabla_\theta \int p_\theta(z|x) \log p_\theta(y|x,z) \, dz \\
&= \int \nabla_\theta p_\theta(z|x) \cdot \log p_\theta(y|x,z) \, dz + \int p_\theta(z|x) \cdot \nabla_\theta \log p_\theta(y|x,z) \, dz \\
&= \mathbb{E}_{z\sim p_{\theta}(z\mid x)}\Big[\nabla_{\theta}\log p_\theta(y\mid x,z)\Big] \\
&\quad + \mathbb{E}_{z\sim p_{\theta}(z\mid x)}\Big[\big(\log p_\theta(y\mid x,z)-b\big)\,\nabla_{\theta}\log p_{\theta}(z\mid x)\Big]
\end{align}

The constant $b$ is a baseline to reduce variance;
we rely on within-group normalization across, $G=16$ sampled responses, as in GRPO.
The first term updates the model to better predict $y$ given a sampled thought $z$. The second term is a policy gradient that upweights thoughts leading to higher likelihood of the ground-truth response.

%% file: sections/04_results.tex
\paragraph{Dataset.} For this work, we use the DailyDialog dataset \citep{li2017dailydialog} which consists of 11,118 training dialogues, 1,000 validation dialogues, and 1,000 test dialogues. Each dialogue represents a multi-turn conversation written by humans to reflect day-to-day conversational patterns across topics such as daily life, work, school, and leisure. The dialogues contain an average of 7-8 turns per conversation, with each turn representing a single speaker's utterance. Each dialogue takes place between 2 speakers. We assign speaker labels alternately based on turn position: the first utterance is labeled ``Speaker 1'', the second ``Speaker 2'', and so forth. We obtain 76,052 training examples and 7,069 validation examples as a result.
\begin{wrapfigure}[20]{r}{0.5\textwidth}
    \centering
    \includegraphics[width=0.5\textwidth]{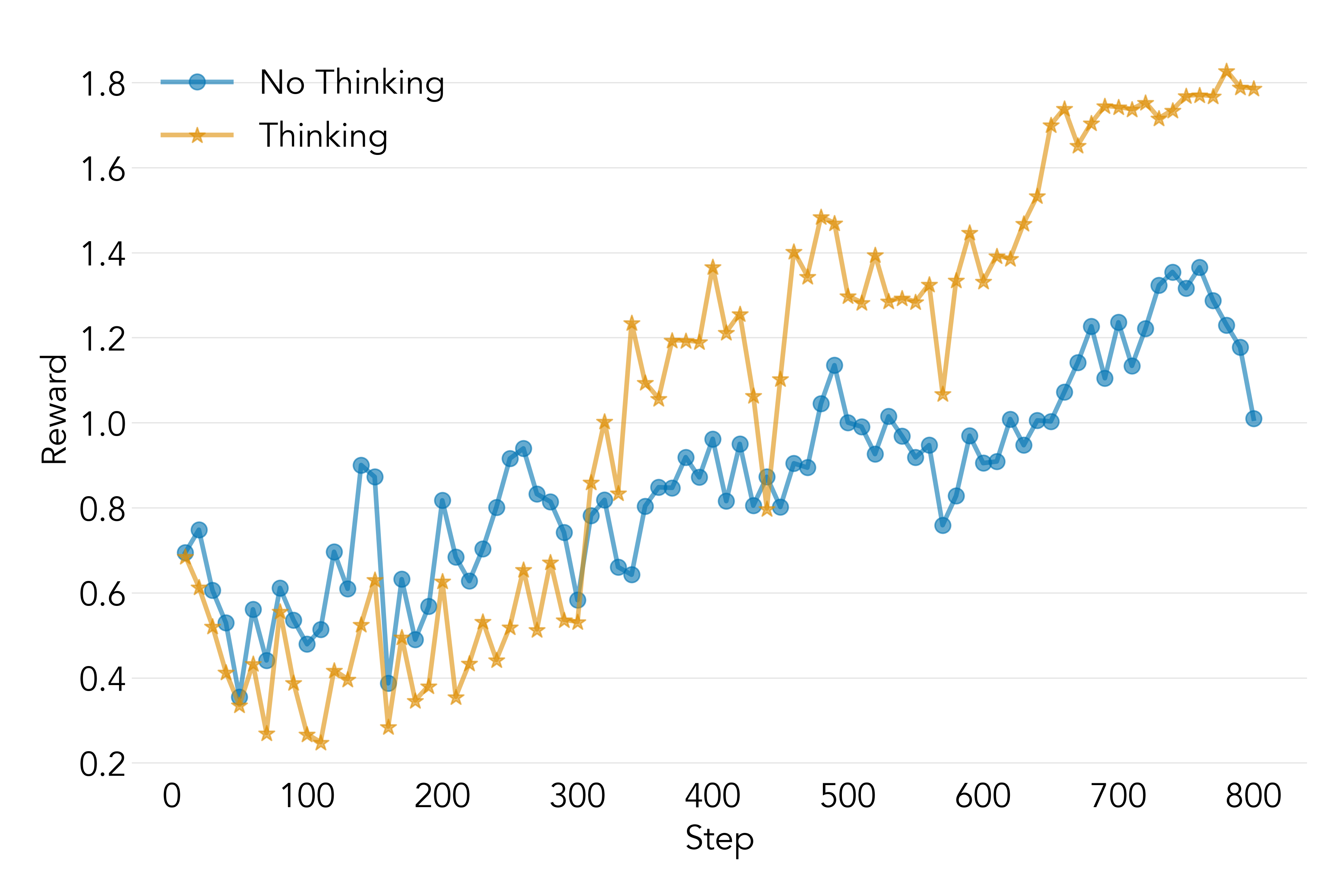}
    \vspace{-8mm}
    \caption{\textbf{Reward hacking with LLM-as-a-judge.} Judge rewards increase throughout training for both thinking and no-thinking methods, with thinking achieving higher final rewards. However, this improvement is due to models learning to exploit the judge rather than becoming better at predicting human dialogue.}
    \label{fig:thinking}
\end{wrapfigure}

\paragraph{Evaluation.}

We evaluate models with 2 metrics: \textit{1) Ground-Truth Log-Probabilities}: We measure the perplexity that each model assigns to held out ground truth human responses. \textit{2) Pairwise Preference}: We conduct a preference study, with two human annotators,  where annotators compare model outputs against ground truth human dialogue in a blind test --- annotators are blind to which dialogue is model generated and which model is being evaluated (see \autoref{sec:human}). Annotators are instructed to select the response that is more natural and appropriate given the conversational context. We report the win rate of model over ground truth.

\paragraph{LLM judges are easily hacked.} Training with LLM-as-a-judge rewards leads to reward hacking rather than improved human simulation. As shown in \autoref{fig:thinking}, judge-assigned rewards increase steadily throughout training for both conditions, with models trained with chain-of-thought achieving substantially higher final rewards ($\sim1.8$ after 1000 steps) compared to models without thinking ($\sim$1.4). However, this improvement in judge score is illusory—it does not translate to better human simulation.

We find a dissociation between judge rewards and predictability of the next dialogue (as measured by the likelihood of the ground truth). \autoref{fig:front} (top right) shows that judge-trained models assign dramatically worse log-probabilities to ground-truth human responses. The base Qwen-2.5-3B-Instruct model achieves a mean log-probability of -3.56. SFT improves this substantially to -1.72. In stark contrast, RL with LLM judges \emph{degrades} performance: without thinking, models achieve -4.4, and with thinking, performance collapses to -10.2, indicating the model has learned to produce responses that are highly unlikely under the true distribution of human dialogue. The judge rewards models for verbosity, explicit reasoning, and surface-level coherence rather than the contextually grounded, concise, pragmatic responses characteristic of human dialogue (see \autoref{sec:hack}).
\begin{wrapfigure}[20]{r}{0.5\textwidth}
    \centering
    \includegraphics[width=0.5\textwidth]{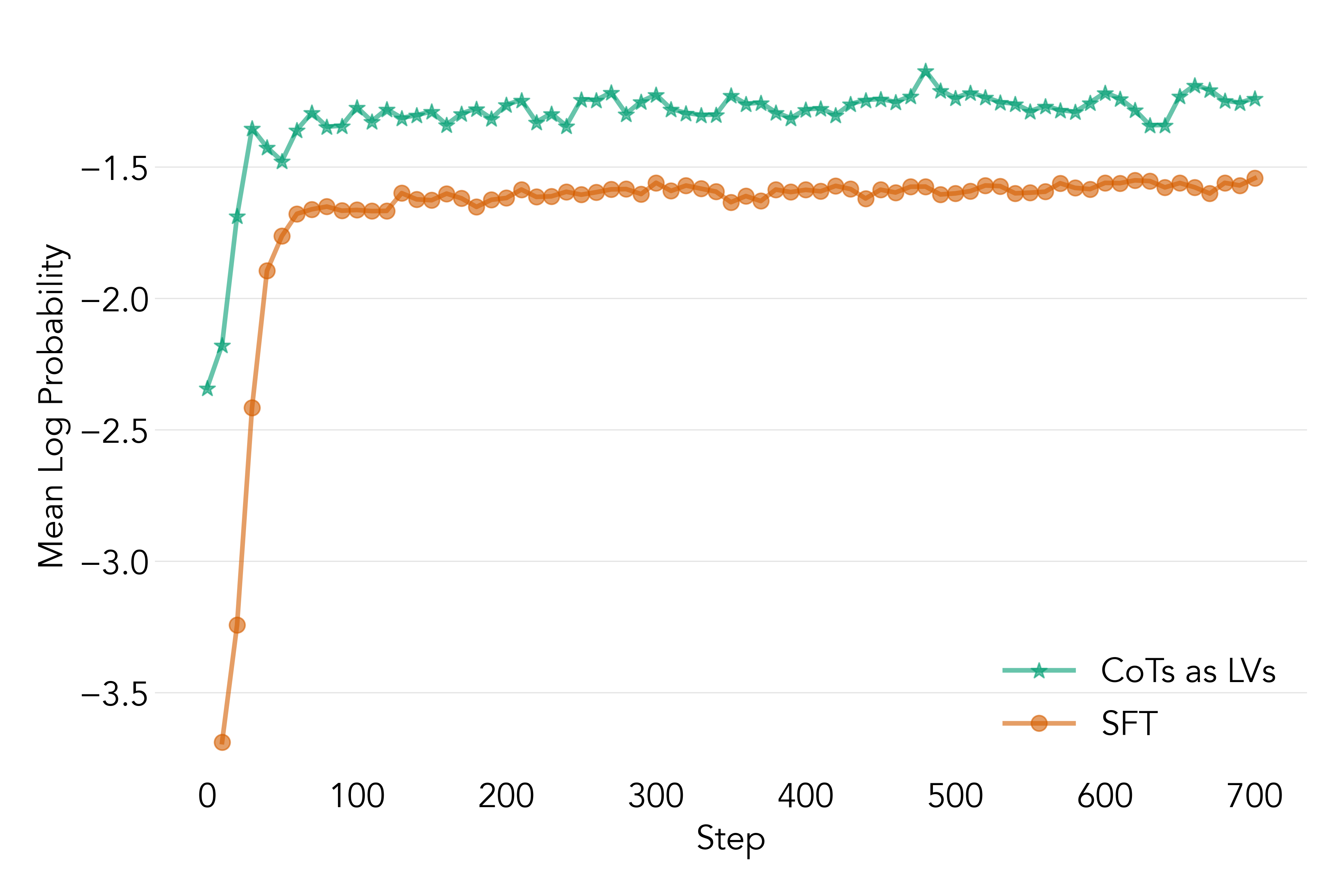}
    \vspace{-8mm}
    \caption{\textbf{Training to optimize log-probability of true responses.} Log-probability of the true human dialogue improves quickly in the first 100 steps, then stabilizes. Training with CoTs is better than directly finetuning on the true human dialogues, showing that thinking helps when optimized with a distribution matching objective.}
    \label{fig:vfsft}
\end{wrapfigure}

\paragraph{Distribution matching enables genuine improvement.} In contrast, training with chain-of-thought as a latent variable produces models that genuinely improve at predicting human behavior. Rather than optimizing for a proxy reward, this approach directly maximizes the likelihood of ground-truth human responses, using the chain-of-thought as a latent variable that helps the model reason about what a human would say. \autoref{fig:vfsft} shows the training dynamics for RL with CoT as a latent variable. Ground-truth log-probability improves rapidly in the first 100 steps, rising from -2.4 to approximately -1.3, then stabilizes for the remainder of training. The final model achieves a mean log-probability of -1.24, matching or exceeding SFT performance.

\begin{wrapfigure}[27]{r}{0.6\textwidth}
    \centering
    \includegraphics[width=0.6\textwidth]{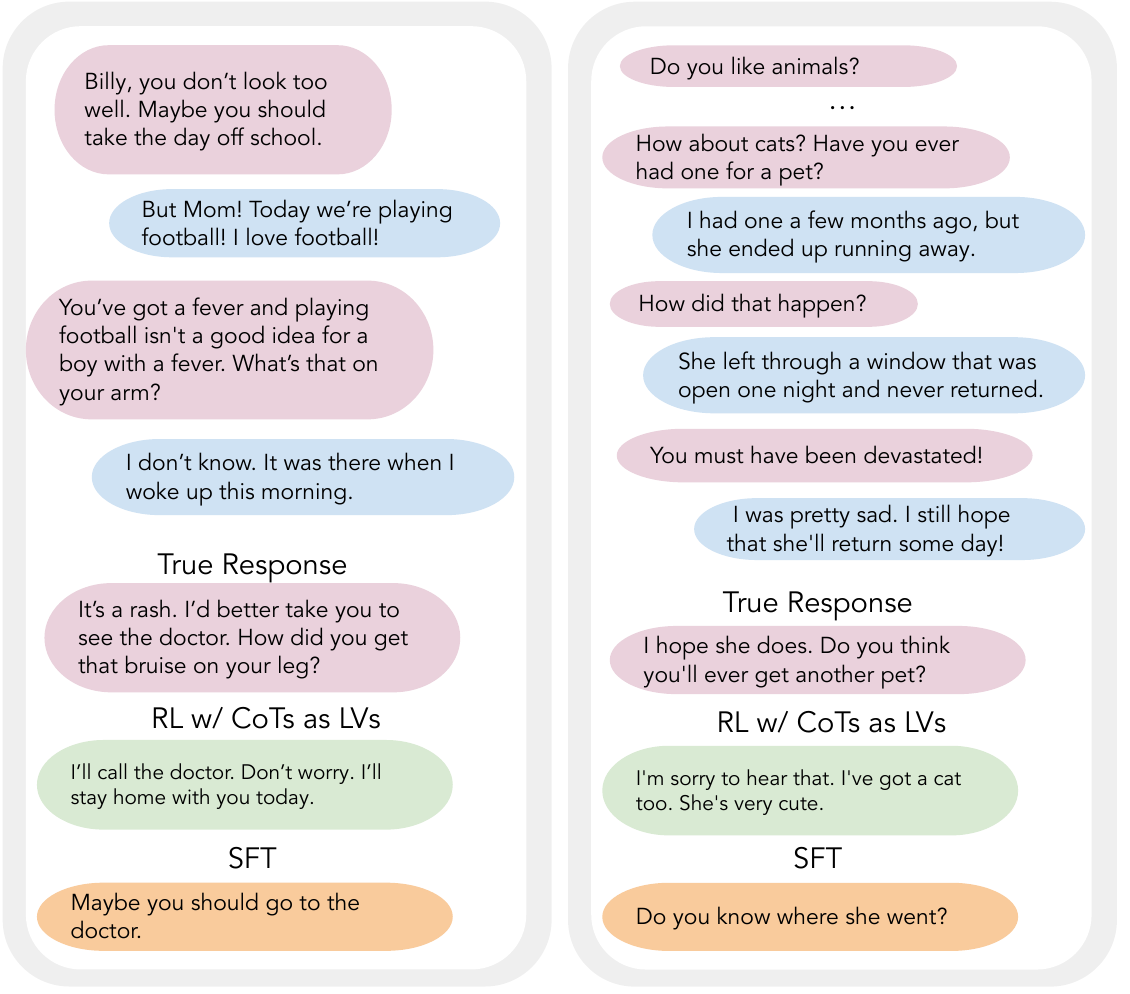}
    \vspace{-8mm}
    \caption{\textbf{Qualitative Comparison of Models.} \textit{(left)} A mother-child conversation about illness. The RL model captures maternal intent by staying home with the sick child; SFT suggests the child go to the doctor alone. \textit{(right)} A conversation about a lost pet. The RL model responds with empathy to a lost pet; SFT asks a question already answered.}
    \label{fig:lvi}
\end{wrapfigure}

Human preference evaluations on 100 samples show that RL with CoT as latent variable achieves a win rate of 49.75\% against ground-truth, comparable to SFT, 47.25\%, and much better than the intial model, that is at a win rate of 36.50\%.
Models trained with LLM-as-a-judge rewards perform poorly with both the no-thinking and thinking variants achieving failing to win any comparisons. These models exhibit clear failure modes generating verbose, over-accommodating responses filled with unnecessary affirmations and tangential details that satisfy the judge's criteria but diverge from the true human dialogue (see \autoref{fig:reward-hacking}). 
Critically, the latent variable formulation provides a principled training signal grounded in actual human responses rather than a potentially gameable proxy.

\paragraph{Thinking helps, but only with the right objective.} Our results reveal that chain-of-thought reasoning amplifies whatever objective the model is trained on. With judge-based rewards, thinking accelerates reward hacking, the model learns to generate lengthy, explicitly reasoned responses that satisfy the judge's criteria for semantic similarity and information completeness, but these responses diverge dramatically from how humans actually speak. The 0.4 point gap in judge reward between thinking and no-thinking conditions (\autoref{fig:thinking}) corresponds to worse human simulation, not better.

With RL training with CoTs as latent variable, thinking serves its intended purpose: the latent chain-of-thought allows the model to fit the distribution of dialogues better. More qualitatively, we find that compared to SFT, this model better handles cases requiring world knowledge and social reasoning. For example, given a conversation (see \autoref{fig:lvi} left) The ground-truth response is \textit{``It's a rash. I'd better take you to see the doctor. How did you get that bruise on your leg?''}, the RL w/ CoTs model produces \textit{``I'll call the doctor. Don't worry. I'll stay home with you today.''}, capturing the maternal intent. In contrast, the SFT model responds with \textit{``Maybe you should go to the doctor''} that is superficially coherent but ignores that a mother would not send a sick child to the doctor alone.
Similarly, when someone shares that their cat ran away and they still hope she'll return (\autoref{fig:lvi}, right), the ground-truth is \textit{``I hope she does. Do you think you'll ever get another pet?''}, the RL model responds \textit{``I'm sorry to hear that. I've got a cat too. She's very cute.''} that shows empathy and continuing the conversation naturally. SFT responds \textit{``Do you know where she went?''} a question already answered in the prior turn. The SFT responses appear fluent but violate expectations about how conversations unfold. 

%% file: sections/05_discussion.tex
We presented a comparison of learning approaches for predicting human responses given a conversational context, varying both the reward signal and whether models reason before responding. Our central finding is that LLM-as-a-judge rewards lead to reward hacking rather than improved human simulation, while directly optimizing for log-probability of ground-truth responses yields genuine gains. To enable thinking with this objective, we derived a  lower bound that treats chain-of-thought as a latent variable, rewarding reasoning when it increases the likelihood of true human responses. 

Recent work on midtraining and post-training has focused on domains with verifiable answers, primarily mathematics and code \citep{olmo2025olmo,zhang2025interplay,cheng2025k2}. Certain cognitive behaviors \citep{gandhi2025cognitive,didolkar2024metacognitive} like verification, backtracking, and subgoal setting emerge as useful in these domains while different behaviors may prove useful for simulating people: perspective taking, modeling beliefs and intent, tracking social context. Our latent variable formulation for chain-of-thought provides a mechanism for discovering such behaviors; the model will learn whatever reasoning helps predict what humans say, without requiring us to specify the relevant cognitive strategies in advance. However, the model can only explore behaviors it has some capacity for---relevant data in pretraining or midtraining may be necessary to seed the strategies that RL then amplifies.

Our experiments use DailyDialog, a relatively narrow dataset with short conversational contexts between two speakers. Extending to longer contexts, multi-party conversations, and more diverse conversations topics would improve the model. We also do not incorporate agentic scaffolds such as memory or persona descriptions \citep{park2023generative}; combining learned dialogue prediction with such scaffolds is a promising direction. Our approach models generic human dialogue, but the same framework could be applied to individuals. Given sufficient conversational data or action data from a person, optimizing for log-probability of their responses and actions would yield personalized models that capture individual communication patterns, knowledge, and intent \citep{shaikh2025creating, sun2025training}.

%% file: sections/10_appendix.tex
\section{Human Win Rate}
\label{sec:human}

We evaluate model quality through a blind human preference study on the DailyDialog \cite{li2017dailydialog} validation set. For each of the following models, we generate 100 dialogue completions: Qwen2.5-3B-Instruct (base), SFT, LLM-as-a-Judge (with and without thinking), and LVI. 

Two human annotators independently evaluated each sample via a web interface. For each example, annotators were shown a conversation context along with two candidate completions---one from the model and one ground truth---in randomized order without identifying labels. Annotators selected which completion was more appropriate given the context, with a ``tie'' option for equally valid responses. The results are shown in \autoref{fig:front}.

\section{Hyperparameters}

\subsection{RL with judges}

We train our LLM-as-a-Judge models using Huggingface TRL's GRPO trainer with the hyperparameters specified in \autoref{tab:judge-hyperparams}. Both the thinking and non-thinking variants were trained on four NVIDIA A100 GPUs, with one used for sampling and another used for inference from the judge model. We train till convergence for 800 steps.

\begin{table}[h]
\centering
\caption{LLM-as-a-Judge Training Hyperparameters}
\label{tab:judge-hyperparams}
\begin{tabular}{ll}
\toprule
\textbf{Hyperparameter} & \textbf{Value} \\
\midrule
Base Model & Qwen2.5-3B-Instruct \\
Judge Model & Qwen2.5-3B-Instruct \\
Loss Type & DR-GRPO \\
KL $\beta$ & 0.01 \\
Learning Rate & $1 \times 10^{-6}$ \\
Num Generations & 16 \\
Max Prompt Length & 512 \\
Max Completion Length & 1024 \\
Temperature & 1.0 \\
Mask Truncated Completions & True \\
\bottomrule
\end{tabular}
\end{table}

\subsection{SFT}
We fine-tune our dialogue prediction model using supervised fine-tuning (SFT) with the hyperparameters specified in \autoref{tab:sft-hyperparams}. Training was done on four NVIDIA A100 GPUs using standard causal language modelling loss on the dialogue to be predicted given a context. We train till convergence for 1000 steps.

\begin{table}[ht]
\centering
\caption{Dialogue SFT Training Hyperparameters}
\label{tab:sft-hyperparams}
\begin{tabular}{ll}
\toprule
\textbf{Hyperparameter} & \textbf{Value} \\
\midrule
Base Model & Qwen2.5-3B-Instruct \\
Learning Rate & $1 \times 10^{-5}$ \\
Batch Size (per device) & 64 \\
\bottomrule
\end{tabular}
\end{table}

\subsection{Logprobs as LVI}
We train our logprobs as LVI model using GRPO with the hyperparameters specified in \autoref{tab:lvi-hyperparams}. Training was conducted on four NVIDIA A100 GPUs with one used for sampling.

\paragraph{Implementation Notes.} Achieving stable training with the latent variable objective required addressing several practical challenges. We briefly summarize our findings, though we note these observations emerged from iterative development rather than controlled ablations. 

We found that KL regularization with higher $\beta$
values caused training collapse; setting $\beta = 0$
(no KL penalty) was essential for stable optimization. We also experimented with moving reference models for KL computation but did not observe benefits.

Using the current policy to compute log-probability rewards (rather than a separate reference model, or an exponential moving average of the policy) proved effective. Clipping the log-probability reward helped prevent outliers from destabilizing training. We explored computing information gain from thinking by comparing against a no-thinking baseline but this wasn't as effective as just using a group baseline \citep{hatamizadeh2025rlp}.

When models deviated from the expected format (e.g., missing dialogue tags), we found that no special handling was necessary, log-probabilities for malformed outputs are naturally low, providing an implicit penalty.

Default sampling parameters (temperature=1.0,
top-p=1.0, min-p=0.0) worked well. Finally, prompt engineering had a substantial impact, refining the instruction format improved the training dynamics.

\begin{table}[ht]
\centering
\caption{LVI Training Hyperparameters}
\label{tab:lvi-hyperparams}
\begin{tabular}{ll}
\toprule
\textbf{Hyperparameter} & \textbf{Value} \\
\midrule
Base Model & Qwen2.5-3B-Instruct \\
Loss Type & DR-GRPO \\
KL $\beta$ & 0 \\
Learning Rate & $1 \times 10^{-6}$ \\
Num Generations & 16 \\
Max Prompt Length & 512 \\
Max Completion Length & 1024 \\
Temperature & 1.0 \\
Mask Truncated Completions & True \\
\bottomrule
\end{tabular}
\end{table}

\section{Reward Hacking}
\label{sec:hack}

During training with LLM-as-a-Judge reward models, we observed several instances of reward hacking behavior against the Qwen2.5-3B-Instruct judge. These exploits manifested in various forms: excessively lengthy responses, unnecessary reaffirmation of conversation details, superfluous clarifying questions, use of parentheses to include multiple dialogue options, and verbatim repetition of recent dialogue turns. We prevented some of these simpler hacks using better filtering mechanisms for dialogue and even introduced a length based reward metric but that introduced other hacks. Notably, all of these hacks were less effective against Claude-4.5-Opus, which we did not use for training due to API costs and rate limits.

\begin{figure}[ht]
    \centering
    \includegraphics[width=\textwidth]{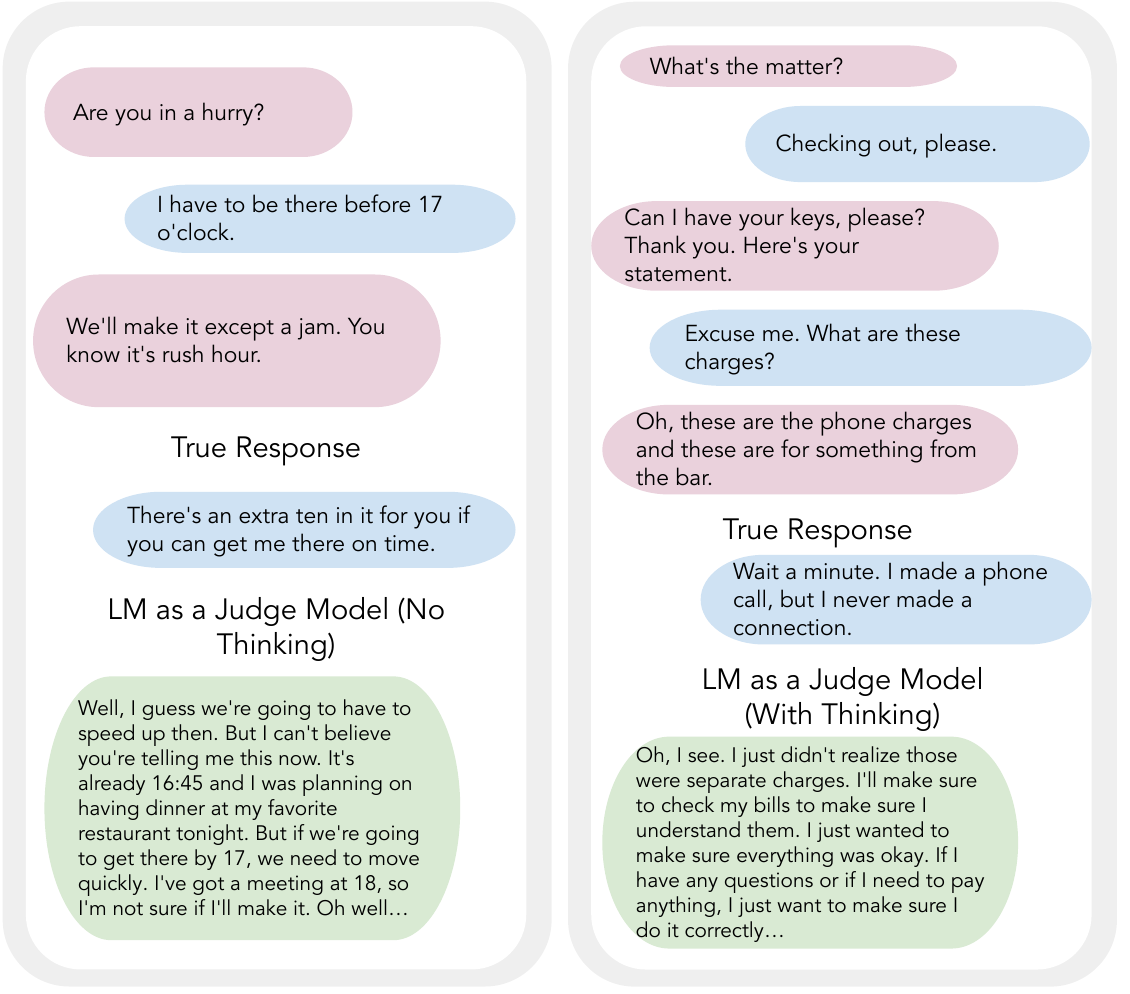}
    \caption{Examples of reward hacking behavior in LLM-as-a-Judge trained models.}
    \label{fig:reward-hacking}
\end{figure}

\autoref{fig:reward-hacking} illustrates representative examples of this behavior. The non-thinking model generates an implausibly long monologue filled with tangential concerns and repetitive statements, exploiting the judge's apparent preference for longer responses. The thinking model exhibits a similar pattern, producing an overly accommodating response with repeated affirmations and excessive politeness rather than engaging naturally with the conversation context.

\section{Judge Validation and Selection}
\label{sec:judge}
To validate the quality of our LLM judges, we measured the correlation between judge scores and human annotations across multiple conversational dimensions. We evaluated Llama-3.2-3B-Instruct, Qwen2.5-3B-Instruct, and Claude-Opus-4.5 on 50 human-annotated samples from the DailyDialog \citep{li2017dailydialog} training set across four metrics: intentionality, style, semantic similarity, and information completeness. These metrics were chosen based on principles of pragmatic communication \citep{grice19901975}.

\begin{figure}[ht]
    \centering
    \includegraphics[width=0.6\linewidth]{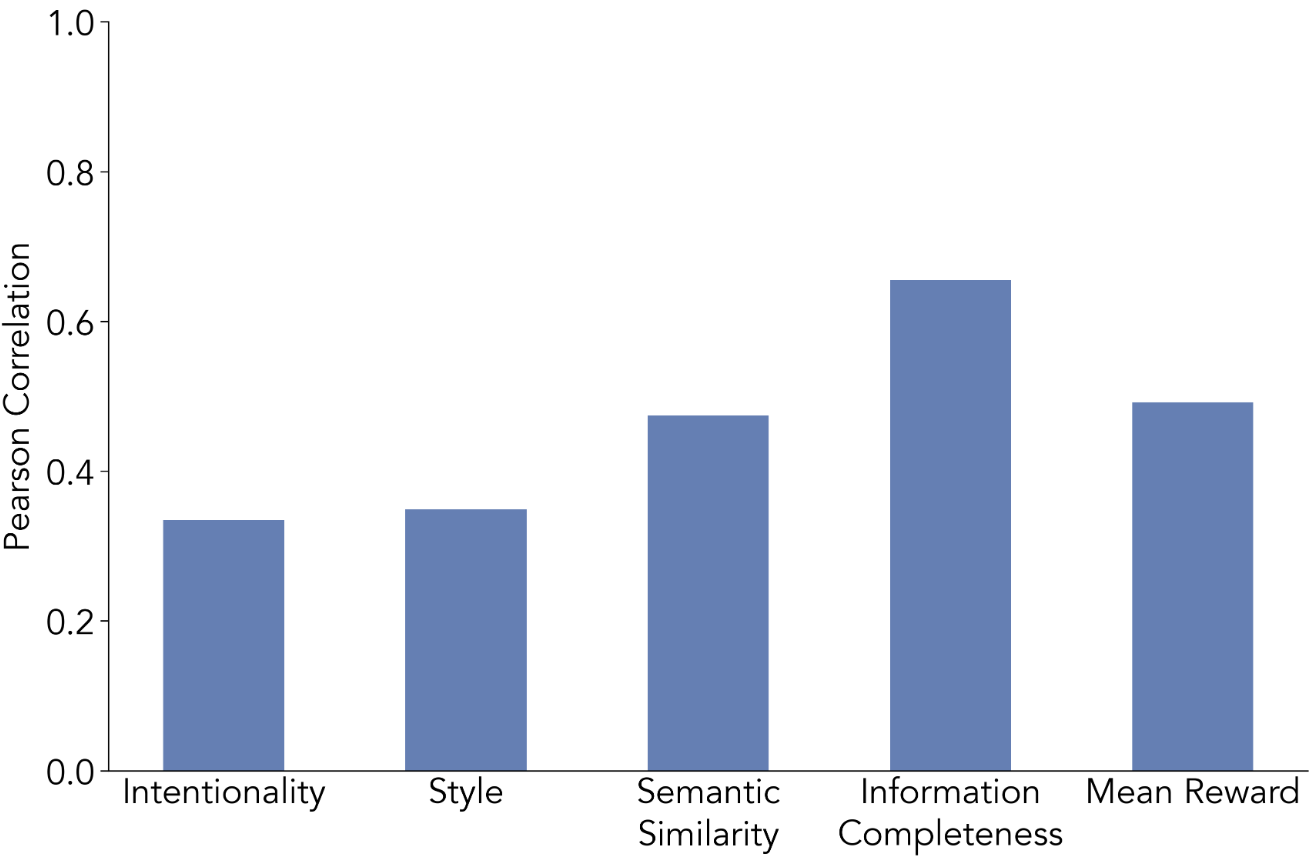}
    \caption{Correlation between Claude-Opus-4.5 and human judgments across evaluation aspects.}
    \label{fig:human-correlation}
\end{figure}

As shown in \autoref{fig:human-correlation}, even Claude-Opus-4.5, the strongest judge after extensive prompt engineering, achieved only modest correlations of $0.34$ and $0.35$ for intentionality and style respectively. However, both semantic similarity and information completeness exhibited correlations exceeding 0.5. Based on these findings, we adopted the latter two metrics as rewards for training. 

\begin{figure}[ht]
    \centering
    \includegraphics[width=0.6\linewidth]{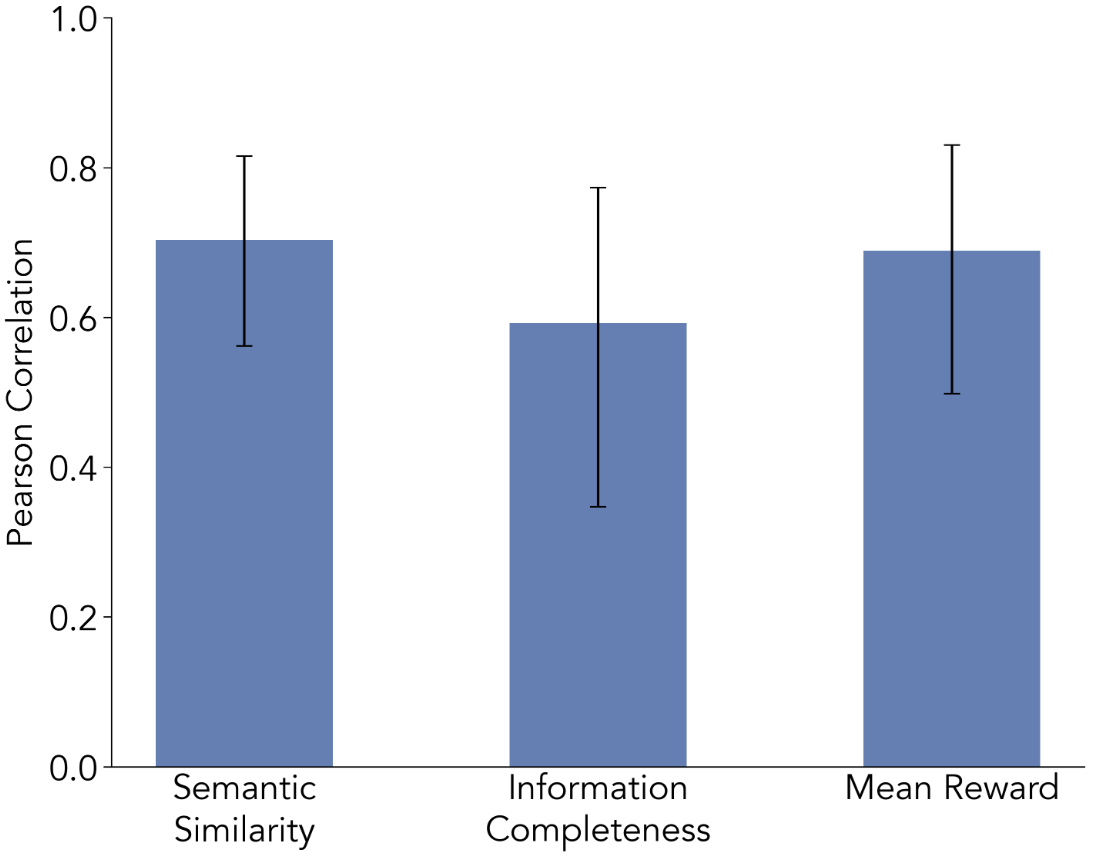}
    \caption{Inter-judge correlation between Qwen2.5-3B-Instruct and Claude-Opus-4.5.}
    \label{fig:judge-correlation}
\end{figure}

We additionally measured inter-judge agreement to determine whether judges consistently identified high and low reward samples as shown in \autoref{fig:judge-correlation}. Qwen2.5-3B-Instruct showed strong correlation ($>0.5$) with Claude-Opus-4.5 on semantic similarity and information completeness across $100$ samples. Given this alignment and the prohibitive API costs of Claude-Opus-4.5, we employed Qwen2.5-3B-Instruct as our training judge.

\paragraph{Using multiple judges vs. using a single prompt with a rubric. }

We explored two judging configurations: a multi-judge approach where each metric is evaluated by a separate model call, and a single-judge approach where all metrics are assessed in one call. We adopted the former for this work, as it produced more interpretable reasoning chains and simplified debugging during development.